\documentclass[letterpaper, 10 pt, conference]{ieeeconf}  

\IEEEoverridecommandlockouts                              

\usepackage{graphicx} 
\usepackage{hyperref}

\title{\LARGE \bf
Deformation of the panoramic sphere into an ellipsoid to induce self-motion in telepresence users
}

\author{Eetu Laukka$^{1}$, Evan G. Center$^{1}$, Timo Ojala$^{1}$, Steven M. LaValle$^{1}$, Matti Pouke$^{1}$
\thanks{This work was supported by Infotech Oulu, the European Research Council Advanced Grant (ERC AdG, ILLUSIVE: Foundations of Perception Engineering, 101020977), and The University of Oulu \& The Research Council of Finland PROFI7 program 352788.}
\thanks{$^{1}$Author is with Faculty of Information Technology and Electrical Engineering,
        University of Oulu, P.O. Box 4500, Oulu, FI-90014, Finland
        {\tt\small firstname.lastname@oulu.fi}}%
}

\begin{document}
© 2025 IEEE. Personal use of this material is permitted.
Permission from IEEE must be obtained for all other uses,
including reprinting/republishing this material for advertising
or promotional purposes, collecting new collected works
for resale or redistribution to servers or lists, or reuse of
any copyrighted component of this work in other works.
This work has been submitted to the IEEE for possible
publication. Copyright may be transferred without notice,
after which this version may no longer be accessible.

This work has been accepted for presentation at the 2025
IEEE Conference on Telepresence, to be held in Leiden,
Netherlands.

\maketitle

\thispagestyle{empty}
\pagestyle{empty}

\begin{abstract}
Mobile telepresence robots allow users to feel present and explore remote environments using technology. Traditionally, these systems are implemented using a camera onboard a mobile robot that can be controlled. Although high-immersion technologies, such as 360-degree cameras, can increase situational awareness and presence, they also introduce significant challenges. Additional processing and bandwidth requirements often result in latencies of up to seconds. The current delay with a 360-degree camera streaming over the internet makes real-time control of these systems difficult. Working with high-latency systems requires some form of assistance to the users.

This study presents a novel way to utilize optical flow to create an illusion of self-motion to the user during the latency period between user sending motion commands to the robot and seeing the actual motion through the 360-camera stream. We find no significant benefit of using the self-motion illusion to performance or accuracy of controlling a telepresence robot with a latency of 500 ms, as measured by the task completion time and collisions into objects. Some evidence is shown that the method might increase virtual reality (VR) sickness, as measured by the simulator sickness questionnaire (SSQ). We conclude that further adjustments are necessary in order to render the method viable. 

\end{abstract}

\section{INTRODUCTION}
Humans can feel present in an environment other than their current physical environment by leveraging immersive technologies. This concept is known as telepresence \cite{Minsky1980Telepresence}. Telepresence often requires a camera that captures the remote location and streams the camera feed to the end-user. Furthermore, if the camera is attached to a mobile platform that the user can control by sending motion commands, the concept expands to robotics and can be called mobile robotic telepresence (MRP) \cite{Kristoffersson2013ATelepresence}. There are different levels of technological immersion in mobile robotic telepresence. Traditionally, we can think of a tablet on wheels when talking about the common social telepresence robot \cite{Kristoffersson2013ATelepresence, Zhang2022TelepresenceReview}. With this type of robot, the user usually uses a computer screen to watch the camera stream, while controlling the robot with either a joystick, keyboard, or pointing toward the goal locations through an interface. However, a more immersive option is using a head-mounted-display (HMD) to render the camera stream from the remote location. The user tends to feel more present in the remote location when the immersion is higher \cite{Cummings2016HowPresence}. Higher immersion can be achieved by using modern technology, such as panoramic or 360-degree cameras, to feel present in another environment than their current physical environment \cite{Jokela2019HowCameras}. 
The 360-degree video is projected onto a spherical surface, and the user's viewpoint is located at the center of that sphere. This enables the user to freely look around the 360-content rendered by the HMD.

\begin{figure}[thpb]
      \centering

      \vspace{-1mm}
      \includegraphics[scale=0.50]{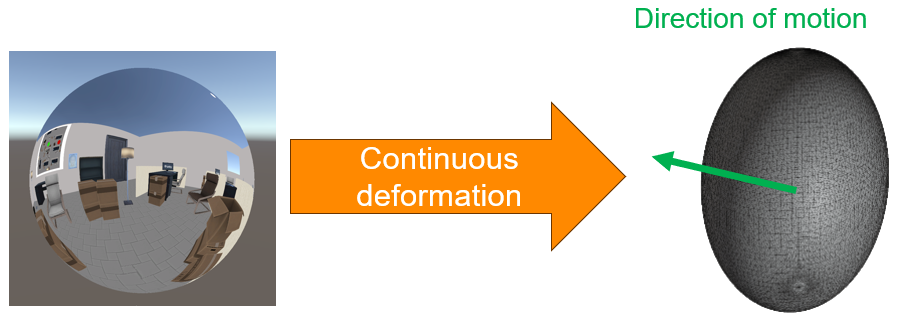}
      \caption{Depiction of how the starting projection sphere is deformed into an ellipsoid with minor axis aligned with the direction of motion.}
      \vspace{-1mm}
      \label{graph}
   \end{figure}
   
\begin{figure}[thpb]
      \centering

      \vspace{3mm}
      \includegraphics[scale=0.45]{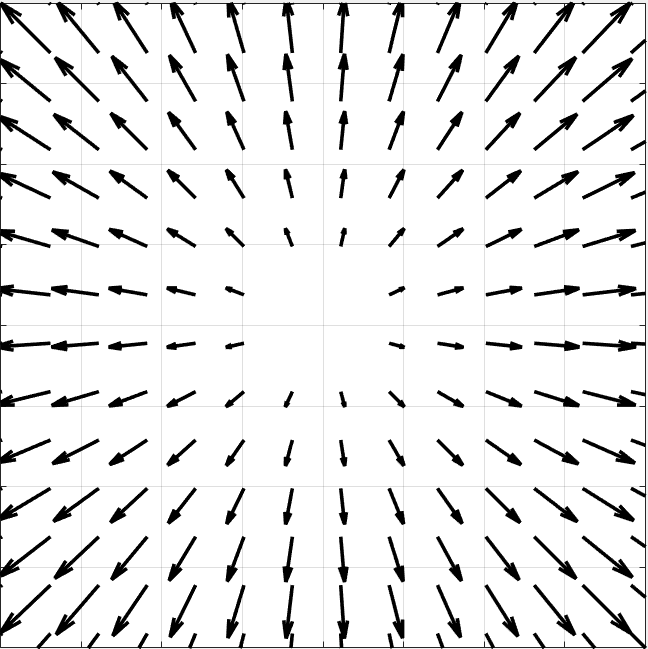}
      \caption{Two-dimensional representation of forward vection prior to its projection onto the retinal photoreceptors. The arrows indicate the motion of visual features.}
      \vspace{-1mm}
      \label{vector}
   \end{figure}

Optical flow is arguably one of the most important sensory inputs that provides the perception of self-motion \cite{Berthoz2002TheMovement}. This flow comes from the moving features that the brain recognizes in the field of view (FOV) of vision \cite{Gibson1950TheWorld}. This can also cause an illusory sense of self-motion called vection, which is a widely studied phenomenon \cite{LaValle2023VirtualReality}. While the general perception of self-motion relies on the integration of multiple sensory modalities, vection is predominantly driven by visual input. There are multiple types of vection that define the direction and type of movement. It is easiest to display them as a 2D rectangular vector field, as in Figure \ref{vector}, which is then projected onto a spherical surface, corresponding to the surface of the photoreceptors on the retina \cite{LaValle2023VirtualReality}. As interactive movement in a 360-degree photo or pre-recorded video is not possible, vection could be leveraged in telepresence systems that utilize 180$^{\circ}$ or 360$^{\circ}$ cameras together with an HMD to view and experience real remote environments. In addition, immersive telepresence systems, where the user can control the position of a remote robot, often suffer from significant latency, making it challenging to effectively control MRP systems \cite{Laukka2024ImprovingVEIPE}. Therefore, it is crucial to explore methods to mitigate the negative effects of latency.

In this paper, we introduce a novel approach in MRP that deforms the rendered panoramic video projection sphere to induce an illusory sense of self-motion. The goal is to give the user immediate feedback on the motions of the robot during the end-to-end latency of the telepresence camera. By gradually transforming the spherical projection into an ellipsoid, we can simulate the optical flow vector field based on the forward motion of the robot before the user sees the actual forward motion of the robot.

\begin{figure*}[t]
      \centering
      \vspace{2mm}
      \includegraphics[width=0.7\textwidth,scale=0.50]{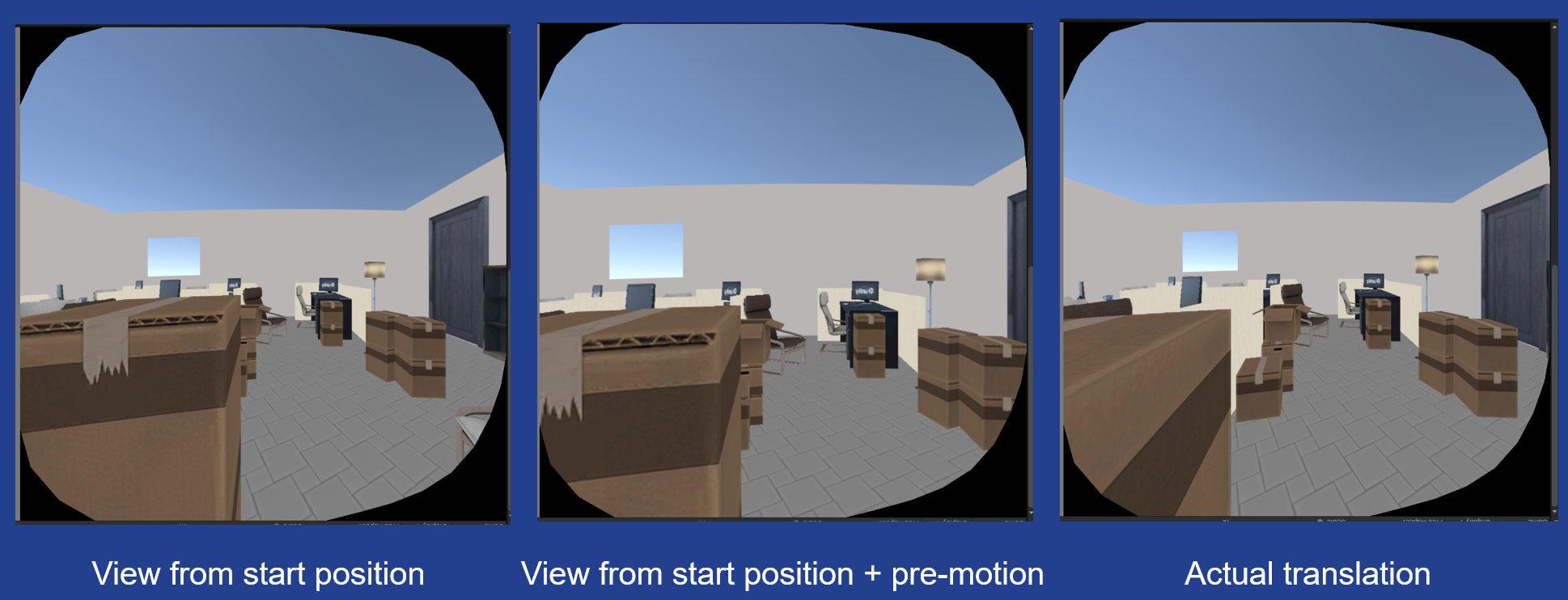}

      \caption {Depiction of the zoom-like effect during the simulated viewpoint translation. \textit{Left}: Initial view from the left eye. \textit{Middle}: Latency phase with ellipsoidal deformation. \textit{Right}: Final view after motion has occured.}
            \vspace{-3mm}
      \label{renderings}
   \end{figure*}

Figure \ref{graph} illustrates this panoramic sphere projection deformation. The left sphere represents the standard projection, where a 360-degree equirectangular image is mapped to the user's viewpoint at the sphere’s center. As the sphere gradually deforms into an ellipsoid, surface pixels stretch to fill the entire field of view of the HMD. This stretching creates a zoom-like effect, making visual features appear to move closer, thereby simulating the optical flow that occurs when moving forward. The zoom-like effect is shown in Figure \ref{renderings}. The left image shows the view from the left eye before any movement occurs. The middle image represents the latency phase, during which the panoramic sphere is temporarily deformed into an ellipsoid to simulate the effect of viewpoint translation. The right image shows the final rendering from the left eye's perspective after the movement has occurred and the panoramic sphere is no longer deformed. By comparing the middle and right images, one can notice that the method is not without limitations. In fact, objects close to the camera seem to not move close enough compared to the real case, and objects far away from the camera seem to move too close or stretch too much. 

Similarly to the technique proposed by \textit{Shimada et al.} \cite{Shimada2022VideoVehicles}, we aim to test the deformation of a panoramic sphere to create an illusion of self-motion in a general immersive MRP. We create a plausible scenario of a user piloting an immersive telepresence robot with a 500 ms latency, tasked to move around a cluttered office to complete simple navigation and observation tasks. A simulated virtual omnidirectional telepresence robot is used to maintain a constant latency and to be able to focus only on the translational latency, as an omnidirectional system does not need to follow the rotation of the user.

Social telepresence robots have traditionally been controlled by moving the remote robot through either a computer interface or a controller that sends motion commands to the robot \cite{Kristoffersson2013ATelepresence}. These robots can have collision avoidance and other semi-autonomous capabilities to help them navigate the remote environment. This type of real-time, interactive, driving-like application has strict latency requirements of around 250 ms \cite{Neumeier2019MeasuringNetworks}. Currently, most social telepresence robots use one or two cameras: one positioned forward for general viewing and another angled downward to aid in navigation \cite{Kristoffersson2013ATelepresence,Almeida2022TelepresenceReview}. This webcam-style configuration has an inherently large drawback of lacking immersion when viewed from a regular monoscopic screen. Higher immersion often leads to a stronger feeling of presence \cite{Cummings2016HowPresence} in the remote environment. In addition, higher immersion tends to yield better perceptual abilities such as situational awareness and depth perception \cite{Martins2009ImmersiveDisplay}.  

Higher immersion can be achieved through streaming a 360-degree view of the remote environment to the user. This requires the use of multiple cameras facing away from the desired viewpoint. The cameras' videos are then stitched together to create the complete panoramic view of the viewpoint \cite{Jokela2019HowCameras}. A pan-tilt camera that can rotate the camera quickly toward the desired orientation can also be used for this purpose \cite{Almeida2022TelepresenceReview}. The clear benefit of 360-degree cameras over pan-tilt cameras is the simplicity of the design, since no added components are needed
to rotate the camera. However, the biggest drawback is the large amount of processing and much higher resolution needed to make 360-degree video viewable satisfactorily on an HMD compared to a traditional screen \cite{Qian2016OptimizingNetworks}.

Current commercially available 360-degree panoramic cameras can introduce delays of up to three to four seconds when streamed over the internet \cite{Yan2022DissectingSystems, Liu2022WhenStreaming}, which is significantly above the 250 ms latency requirement for real-time applications \cite{Neumeier2019MeasuringNetworks}. When delays exceed this threshold, some other form of support should be given to the user. One proposed solution is some form of predictive display that provides the user a prediction of their actions by taking into account the delay and input commands \cite{Sheridan1993SpacePrognosis}. This prediction is then shown to the user by some indicator, for example, by showing an estimate of where the robot has already moved based on the input commands. Another option is to let the user use a virtual copy of the remote physical room as an interface, to control the robot movements \cite{Laukka2024ImprovingVEIPE}.

A previous study tested the deformation of the panoramic sphere in a simple setting. \textit{Shimada et al.} managed to evoke pre-motion to the users by deforming the projection sphere into an ellipsoid to help users onboard self-driving cars to reduce the perceived acceleration 
\cite{Shimada2022VideoVehicles}.

In a study by \textit{Luo et al.} the researchers managed to utilize a similar zoomed version of the view to increase translational gain thresholds. However, they observed that users are more prone to VR sickness, when the view is zoomed while navigating, compared to a regular view \cite{Luo2024WalkingGain}.

In virtual environments (VEs), optical flow has been shown to be used to manipulate self-motion perception, especially when the manipulation is done in the periphery view \cite{Bruder2011TuningIllusions}. Moreover, the perception of speed can be affected by utilizing optical flow to create the sensation that the user is moving faster or slower than the actual movement \cite{Okuno2008ASensation}. A study by \textit{Shimada et al.} utilized the deformation of the panoramic sphere to successfully induce pre-motion to users in order to lower the perceived acceleration of users aboard self-driving cars, by using the technique to warp the panoramic projection sphere into an ellipsoid to simulate the optical flow of moving forward \cite{Shimada2022VideoVehicles}. Instead of only influencing the perception of acceleration, we aim to use this method to generate a sense of self-motion during the latency between control and visual feedback of the MRP. While \textit{Shimada et al.} focus on a specific application in autonomous vehicles, the present work adopts a general telepresence framework.

\section{METHODS}
The goal of the study was to utilize optical flow to create an illusion of self-motion in order to increase the accuracy of controlling the telepresence robot in a high-latency environment and to mitigate sickness by masking the delay between the action and the perception of the action through the video stream. The aim was to simulate the regular optical flow of a forward-moving camera by gradually distorting the 360-degree panoramic rendering sphere into an ellipsoid to achieve a similar effect. This was designed to give the user immediate feedback of their translational actions during the round-trip delay between the action and seeing that action's effect through the 360-degree camera video stream.

We constructed a pilot experiment in order to test this method and estimate how large of a sample would be required to detect differences in performance, as measured by the time required to complete a navigation and observation task. Moreover, we wanted to measure the effect on VR sickness, by using the SSQ total score from the SSQ questionnaire \cite{Kennedy1993SimulatorSickness}. The simulated office environment featured tight corners and many boxes, allowing an opportunity to calculate the number of collisions into walls and objects as a measure of the navigation accuracy of the participants.

\subsection{Participants}
We collected a sample of 20 participants with the goal of conducting a rigorous power analysis on the effects of interest. The sample was gathered using the University of Oulu Sona system, which consists of participants from the wider university community. Seven out of the 20 were female and 13 were male. The mean and standard deviation of the sample age was $M = 29.6, sd = 5.28$. The participants were rewarded with a small compensation of University of Oulu merchandise worth around 10 euros. All the participants reported to have either normal or corrected-to-normal vision. Our protocols were approved by the University of Oulu's ethical review board.




\subsection{Procedure}
The experiment was conducted using a within-subjects design. There were two conditions: a control condition, which acted as a baseline telepresence system, where the simulated room was captured using a virtual 360-degree camera and rendered with an added artificial latency of 500 ms to the panoramic sphere around the HMD viewpoint; and an experimental condition, which used the gradual distortion of the panoramic sphere into an ellipsoid to induce self-motion illusion to the participants for the duration of the latency. The experimental condition ensured that when participants issued a forward command, they received immediate visual feedback simulating camera movement, prior to the arrival of the actual camera feed from the robot.

Current commercial 360-degree cameras often experience streaming latencies of up to three seconds when transmitting over the internet \cite{Yan2022DissectingSystems, Liu2022WhenStreaming}, which is far too long for effective real-time control of telepresence robots. To address this, we selected a lower latency, which was still twice the typical requirement for real-time navigation \cite{Neumeier2019MeasuringNetworks}. Additionally, our goal was to apply the sphere deformation subtly, trying to ensure that users would not perceive any manipulation. Excessive deformation of the spherical view introduces noticeable distortions.

The participants' task in both conditions was the same; they had four positions in the virtual office to which they needed to navigate and say aloud either a numbered code at the position or look into a bowl and tell the color of the ball inside. The codes and the colors of the balls changed between the conditions. In addition, the order of the navigation goals differed between the two conditions, leading to different paths to mitigate learning effects. The next target position was visible to the participants on a large map on the wall of the office, which was visible from all locations. The participants were instructed that it was not possible to go through objects, but were not instructed specifically to avoid collisions. However, when the simulated telepresence robot was touching any object, it caused the camera to shake which was designed to make the users aware of collisions and naturally to try to avoid them.

Following each condition, participants completed a questionnaire consisting of the SSQ questionnaire and the NASA Task Load Index (NASA-TLX) \cite{Hart1988DevelopmentResearch}. They were also asked to identify which of the two task sessions they found easier and to provide open-ended feedback explaining their reasoning.

\subsection{Apparatus}
\begin{figure}[htpb]
      \centering

      \vspace{3mm}
      \includegraphics[scale=0.15]{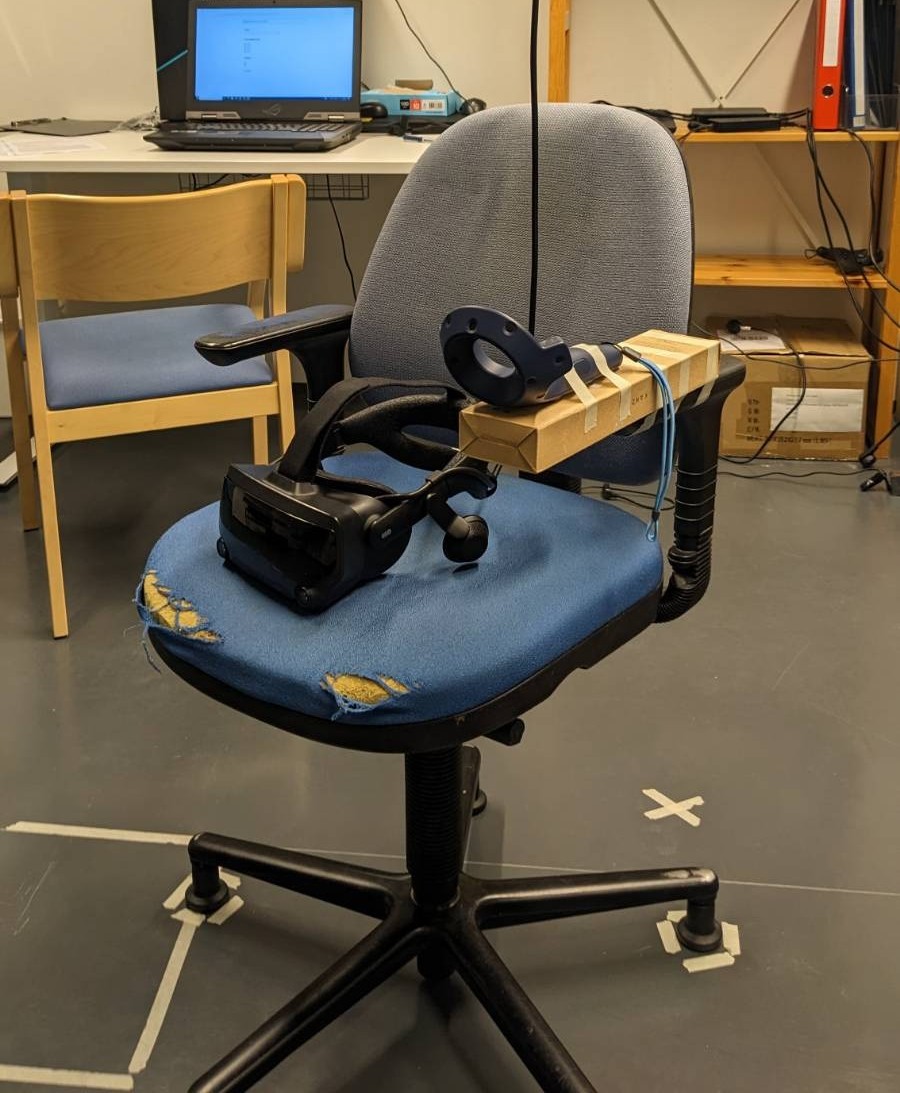}
      \caption{Experimental setup with a rotating office chair equipped with a tracker to track the chair heading during rotation.}
      \label{chair}
   \end{figure}

The simulated telepresence robot was capable of moving omnidirectionally. The participants could control the heading of the robot by rotating the office chair while seated, as depicted in Figure \ref{chair}. In addition, they could move forward and backward by pushing the Valve Index controller joystick forward or backward. The participants used their dominant hand to control the translational movements. The HTC Vive controller mounted on the office chair was used to track its rotation, while a Valve Index headset served as the HMD during the study.

The virtual environment was designed in Unity 2021.3.15f1 Game Engine. The virtual environment was a simple, cluttered office, where the participants could navigate freely with the simulated immersive telepresence robot. One view of the virtual environment is shown in Figure \ref{renderings}.

\subsection{Analyses}
The hypotheses were evaluated based on the performance metric of the participants and the number of detected collisions. VR sickness was identified as a potential risk associated with the method due to distortions in the panoramic rendering sphere. We measured VR sickness using the SSQ questionnaire and assessed perceived workload using the NASA-TLX questionnaire.

Task completion time and collisions were recorded using the Unity Game Engine, which logged data on every frame of the game engine during the experiment. This allowed for a precise analysis of the number of times participants collided with walls or objects and the total time taken to complete the tasks.

To test for significant differences between conditions, we used the Wilcoxon signed-rank test for matched pairs. Furthermore, participants answered a forced-choice question: ``Remembering the two times you did tasks, which one of them felt easier to complete?'' to compare perceived task difficulty. The responses were analyzed using an exact binomial test.

Finally, open-ended feedback was collected as an informal measure of the opinions of the participants and to collect suggestions to improve the method.
%


\section{RESULTS}

\subsection{VR sickness}
\begin{figure}[thpb]\vspace{2mm}  
      \centering
      \includegraphics[scale=0.35]{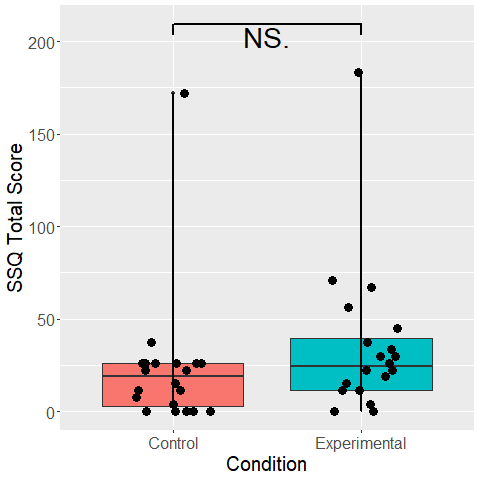}
      \caption{Boxplots comparing the SSQ total score distributions between the two conditions.}
      \vspace{-5mm}      
      \label{SSQfigure}
   \end{figure}

We hypothesized that VR sickness would be significantly reduced under experimental conditions. However, as shown in Figure \ref{SSQfigure}, no significant reduction in the SSQ total score was observed. In fact, when using the Wilcoxon signed-rank (two-sided, alpha = .05) test for matched pairs, the control condition showed a strong trend towards lower SSQ Total Scores ($Mdn = 18.7, sd = 37.05$) compared to the experimental condition ($Mdn = 24.31, sd = 40.73$), $Z = -1.43, p = 0.076,  r = 0.32$.
%
%
%
%
%
%
%
%

%
%

\subsection{Performance}
\begin{figure}[thpb]\vspace{2mm}  
      \centering
      \includegraphics[scale=0.45]{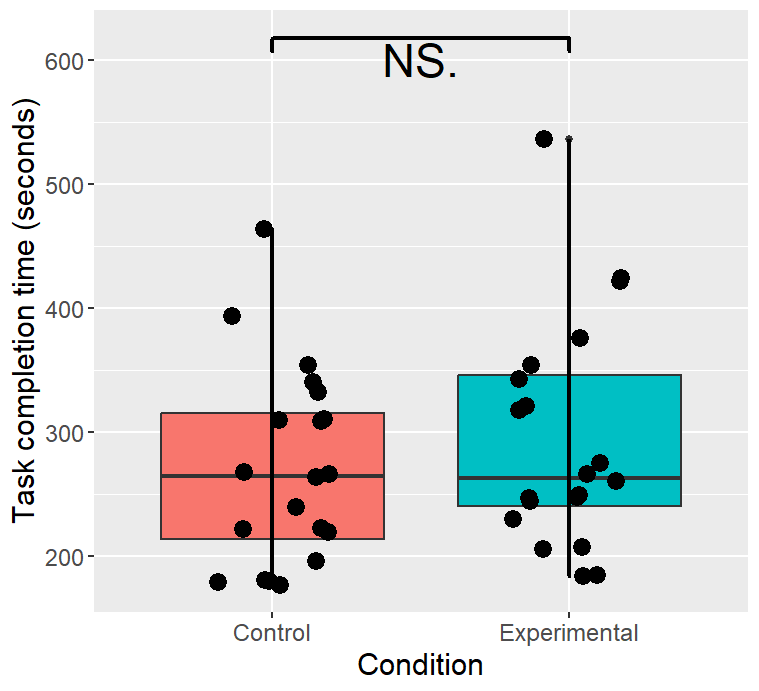}
      \caption{Boxplots comparing the total time distributions in seconds to complete all tasks between the two conditions.}
      \vspace{-5mm}      
      \label{time}
   \end{figure}

Participants completed four tasks in counterbalanced order, each following a path of similar length. The total time to complete all four tasks was recorded and analyzed using a Wilcoxon signed-rank (two-sided, alpha = .05) test. The results did not indicate a significant improvement in task completion time for the experimental condition ($Mdn = 263.13, sd = 91.93$) compared to the control condition ($Mdn = 265.01, sd = 79.47$), $Z = -0.88, p = 0.19,  r = 0.20$. Figure \ref{time} presents the performance results.

\subsection{Collisions}
\begin{figure}[thpb]\vspace{2mm}  
      \centering
      \includegraphics[scale=0.35]{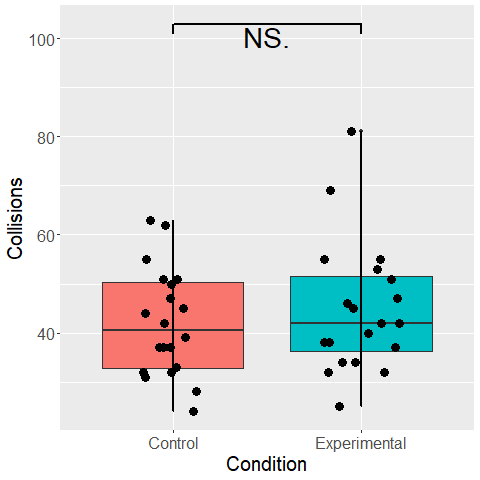}
      \caption{Boxplots comparing the distributions of total collisions between the two conditions.}
      \vspace{-5mm}      
      \label{collision}
   \end{figure}

When a participant collided into anything, a collision event was recorded. Continuously pushing into an object did not cause the object to move, but caused shaking of the simulated camera. We predicted that with the added self-motion simulation, the participants would be more accurate with their controls. However, we compared the results using the Wilcoxon signed-rank (twosided, alpha = .05) test for matched pairs and did not find a significant difference in total collisions from the control condition ($Mdn = 40.5, sd = 10.95$) to the experimental condition ($Mdn = 42, sd = 13.30$), $Z = -0.08, p = 0.47,  r = 0.017$. Figure \ref{collision} presents the collision results.

\subsection{NASA-TLX and user feedback}
The participants filled the NASA-TLX questionnaire after each condition. We calculated the Raw-TLX \cite{Hart2006NASA-taskLater} scores and compared them using the Wilcoxon signed-rank (two-sided, alpha = .05) test. In Raw-TLX, each component of the NASA-TLX questionnaire are treated equally. The results showed no significant difference in the comparison of the average value of the components (p \textgreater .05).

The participants were also asked: ``Remembering the two times you did tasks, which one of them felt easier to complete?'' as a forced choice question and they could answer openly on why they chose each answer. Only 6 out of the 20 chose the experimental condition. Anecdotally, some participants reported that they felt lower latency or no latency in the experimental condition.

\section{DISCUSSION}
Our results suggest that the gradual distortion of the panoramic sphere can be used to create an illusory sensation of self-motion like in the research by \textit{Shimada et al.} \cite{Shimada2022VideoVehicles}. Whereas their study focused on acceleration perception, we present some evidence suggesting the ability to evoke a more general sense of self-motion in a telepresence framework. However, more pertinently, the results suggest that the method can cause an increase in VR sickness as the trend of the SSQ total score seems to rise from the control condition to the experimental condition. This is likely to be due to the distortions caused by the deformation of the rendering sphere, as it was mentioned in informal feedback by some users. The distortions were strongest on the sides of the panoramic sphere. This may explain why the aforementioned study reported no differences in comfort, as participants were only looking ahead and the deformation period seemed to be very brief. Moreover, when the panoramic sphere is deformed and therefore the environment is in a more zoomed-in view for the participant, unnatural visual motions can be the cause of the users getting VR sick, giving some support to the findings by \textit{Luo et al.} \cite{Luo2024WalkingGain}.

While \textit{Shimada et al.} \cite{Shimada2022VideoVehicles} reported no significant difference in comfort between the two experimental conditions, both \textit{Luo et al.} \cite{Luo2024WalkingGain} and the present study observed increased sickness. This suggests that perceived comfort may be case-specific. Further clarification could be achieved through more in-depth analysis, on the one hand probing the methods used to simulate translations of the viewpoint, and on the other hand investigating the relationship between specific application contexts and sickness.

For this study, we conclude that the technique has potential, but needs to be restricted so that the users cannot observe the nauseating distortions happening on the sides of the panoramic sphere. For example, eye tracking could be used so that the self-motion illusion is only used when the participants are looking to the direction of the forward motion. Since immersive technologies should focus on making systems that would not cause VR sickness, we decided not to conduct a confirmatory study, since the original technique would need to be drastically changed. These results serve as an informative guide towards designing novel self-motion illusions for telepresence systems with high latency.

Furthermore, we did not find evidence that the performance or control accuracy of the users would benefit from the induced self-motion during the video stream latency. An increase in VR sickness can have an effect on task performance \cite{Mimnaugh2023VirtualExperiences}, which can partly explain why participants did not perform better. Moreover, the latency (500 ms) was not significantly higher than the usual requirement of 250 ms. This could mean that the participants were able to control the robot fairly well in this latency magnitude without the induced self-motion giving any assistance. Higher latency could make robot control more difficult, and in this case, deformation as an assistive method could lead to a larger increase in performance. However, it should be kept in mind that VR sickness still needs to be kept as low as possible. Finally, participants were unable to avoid obstacles better with the experimental condition. This could stem from the collisions not being punishing enough by only shaking the camera when a collision happens. The participants were observed to not care at all about the collisions, and that could mask any beneficial effect on the performance, as the users did not care regardless.
\vspace{-1.5mm}
\subsection{Limitations}
\vspace{-1mm}
The experiment was a pilot study to assess the usefulness of the proposed method. The results were suggestive, but not conclusive, and did not justify conducting a follow-up confirmatory study. 
However, the findings can be used as an informative guide to build upon future work and refine the method with the objective of researching new ways to use self-motion illusion to manipulate users perception without a rise in the experienced VR sickness. In addition, the experiment used a simulated 360-degree camera in a virtual environment and future work should focus on testing the method in a real immersive telepresence system as the latencies can vary between a large range. The latency in our experiment was kept constant, which does not reflect the variability typically encountered in real-world applications. In practical scenarios, sudden spikes in latency could cause the deformation to exceed comfortable levels, resulting in noticeable and potentially disruptive visual distortions. Adapting the deformation to the latency could help mitigate this problem.
\vspace{-1.5mm}
\subsection{Future Work}
\vspace{-1mm}
Future studies should either restrict users’ head rotations or incorporate eye tracking to apply the method only when participants are looking in the intended direction. This approach is based on our hypothesis that distortions during head movement contribute to increased VR sickness. This ensures that the simulated forward motion optical flow stays as natural as possible. Moreover, simulated auditory cues should be considered since this can make the perceived self-motion illusion stronger as additional sensory modalities has been shown to cause more convincing sensations of locomotion in VR \cite{Moullec2024ToPerspectives}. The method should be tested on higher latencies of up to 3 seconds, as current 360-degree camera streams over the internet can experience delays of that magnitude. Finally, collisions should be penalized more heavily to discourage participants from making contact with objects, as such impacts could pose safety risks in real-world applications.

\addtolength{\textheight}{0 cm}   




\vspace{-1.5mm}
\section*{ACKNOWLEDGMENT}

We would like to thank all our participants for supporting this research.

\bibliographystyle{ieeetr}

\bibliography{root}


\end{document}